\documentclass[pdflatex,sn-basic,Numbered]{sn-jnl}
\usepackage{graphicx}
\usepackage{multirow}
\usepackage{amsmath,amssymb,amsfonts}
\usepackage{booktabs}
\usepackage{enumitem}
\usepackage{microtype}
\usepackage{etoolbox}
\setlist[itemize]{leftmargin=1.5em}
\setlist[enumerate]{leftmargin=1.5em}
\makeatletter
\patchcmd{\@maketitle}{\removelastskip\vskip24pt}{\removelastskip\vskip10pt}{}{}
\patchcmd{\@maketitle}{\removelastskip\vskip24pt}{\removelastskip\vskip10pt}{}{}
\patchcmd{\@maketitle}{\removelastskip\vskip20pt}{\removelastskip\vskip16pt}{}{}
\makeatother

\begin{document}
\title[VTG-RAG for Long-Video Reasoning]{VTG-RAG: Long-Video Reasoning via Query-Adaptive Coarse-to-Fine Sampling and Temporal Graph Retrieval}

\author[1,2]{\fnm{Yao} \sur{Lu}}
\author[1,2]{\fnm{Yunhao} \sur{Wang}}
\author[1,2]{\fnm{Yaxin} \sur{Gao}}
\author[1,2]{\fnm{Feifei} \sur{Teng}}
\author[1,2]{\fnm{Chenxiao} \sur{Liu}}
\author[2]{\fnm{Zhe} \sur{Tang}}
\author[3]{\fnm{Jiaheng} \sur{Wei}}
\author[2]{\fnm{Zhaowei} \sur{Zhu}}
\author*[1,2]{\fnm{Shanqing} \sur{Yu}}\email{yushanqing@zjut.edu.cn}
\author[1,2]{\fnm{Qi} \sur{Xuan}}

\affil[1]{\orgdiv{Institute of Cyberspace Security}, \orgname{Zhejiang University of Technology}}
\affil[2]{\orgdiv{Binjiang Institute of Artificial Intelligence}, \orgname{Zhejiang University of Technology}}
\affil[3]{\orgname{Hong Kong University of Science and Technology (Guangzhou)}}

\abstract{Long-video reasoning requires a system to localize sparse evidence over extended temporal sequences, preserve temporal structure, and generate answers despite substantial multimodal distractors. Existing video retrieval-augmented generation methods typically build indexes from uniformly sampled video segments. In long-video settings, this design readily introduces irrelevant context and is unreliable for questions with explicit temporal constraints. We propose VTG-RAG, a framework for long-video question answering that combines query-adaptive coarse-to-fine sampling with temporal graph retrieval. First, a two-stage sampling strategy selects core segments with high mean query relevance or high intra-segment relevance variance. The framework then integrates subtitles, speech transcripts, visual descriptions, entities, and relations from these segments into a temporally ordered multimodal graph. Finally, it performs temporally filtered, two-pass retrieval and reasoning across frame, text, and graph evidence spaces. We evaluate VTG-RAG on Video-MME and LongVideoBench. On LongVideoBench, the reported Qwen2.5-VL-32B and 72B results improve from 48.16\% to 50.56\% and from 52.13\% to 53.32\%, respectively, with 24 answer-input frames. These correspond to gains of 2.40 and
1.19 percentage points. Ablation experiments on Video-MME support the contributions of query-adaptive sampling, graph retrieval, and two-pass reasoning.}
\maketitle

\section{Introduction}
Long video reasoning is becoming an essential capability in the field of visual language models, supporting applications ranging from movie and sports analysis to surveillance and egocentric assistance~\cite{zhou2025mlvu,wang2025lvbench}. Unlike static images or short clips, long videos distribute visual and linguistic evidence across extended timelines, such that answering a question may require locating a brief event and relating it to observations occurring far earlier or later~\cite{ren2024timechat}. Directly processing all frames is often impractical because long video sequences can exceed the context capacity of current multimodal large language models and incur substantial computational and memory costs~\cite{shu2025videoxl,shen2025longvu}. Meanwhile, aggressive visual compression may discard fine-grained yet task-critical cues~\cite{qian2024streaming,zhang2025xcomp}. Long-video reasoning therefore requires more than simply extending the context window: a system must localize sparse question-relevant evidence, preserve temporal relationships, and integrate information across distant segments into a compact representation for reliable reasoning~\cite{wang2025videotree,hannan2025revisionllm}.

Recent long-context vision-language models and video retrieval-augmented generation methods have substantially advanced long-video understanding~\cite{ren2025videorag,song2023moviechat,song2024moviechatplus,zhang2024longva}. Nevertheless, long-video understanding and moment-retrieval pipelines commonly begin with uniformly sampled frames or fixed-length clips before subsequent aggregation, retrieval, or reasoning~\cite{afham2023kts,fang2024spotvmr}. This strategy treats sampling as a fixed preprocessing step and allocates similar computational resources to temporal regions with substantially different relevance. Increasing the number of sampled frames may reduce the risk of missing evidence, but it also expands the context and introduces additional distractors. Conversely, using fewer uniformly sampled frames may discard decisive evidence before indexing, which downstream retrievers and answer models cannot recover~\cite{hu2025frameselection}. This limitation cannot be fully addressed by selecting segments solely according to their average similarity to the question. Even when a segment appears only moderately relevant as a whole, it may contain a few question-critical frames whose sparse signals are overwhelmed by the surrounding irrelevant content~\cite{li2023transtr,liu2025bolt}. Moreover, when uniformly sampled segments are directly converted into multimodal chunks or graph nodes, irrelevant content propagates into the retrieval structure and reduces its information density. Effective long-video reasoning therefore requires sampling to be optimized jointly with downstream evidence organization. Existing sampling compression methods reduce redundant visual input~\cite{zemskova2026focusgraph,zhu2025focus}, while graph-based retrieval methods organize dispersed evidence for multi-hop reasoning~\cite{edge2024graphrag,guo2024lightrag,zhu2025kg2rag}. However, these components are typically studied separately, leaving unresolved how question-dependent sampling should determine the compact, temporally structured evidence used by downstream retrieval and reasoning.

\begin{figure}[t]
  \centering
  \includegraphics[width=\textwidth]{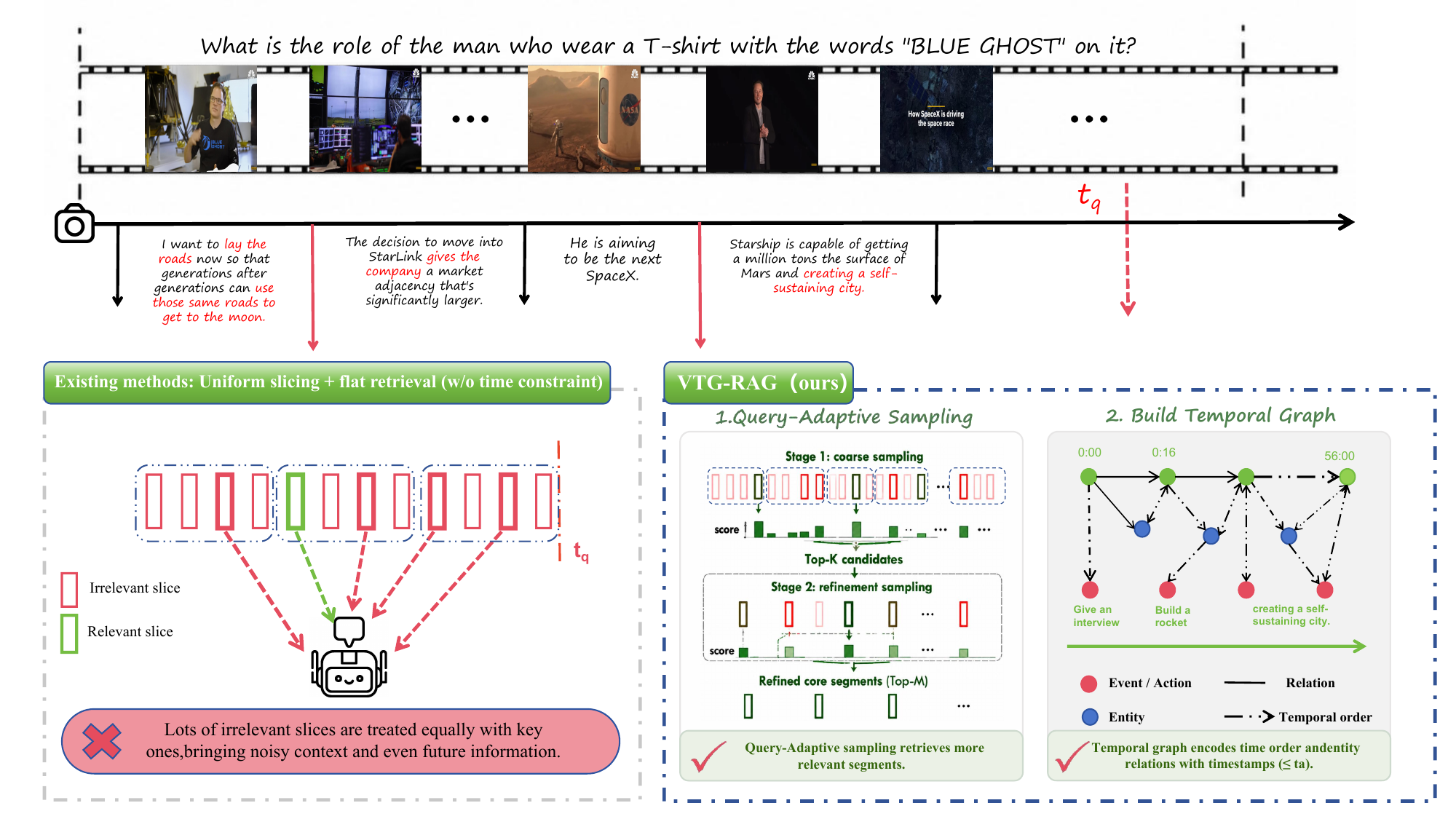}
  \caption{Motivation for VTG-RAG. Uniform segmentation treats all temporal intervals equally, introducing irrelevant or temporally invalid evidence into long-video question answering. In contrast, VTG-RAG combines query-adaptive coarse-to-fine sampling with temporal graph retrieval to construct a compact and ordered evidence context, applying timestamp-based filtering when a query time is available}
  \label{fig:motivation}
\end{figure}

To address this gap, we propose VTG-RAG, which consists of query-adaptive coarse-to-fine sampling with temporal graph retrieval. As illustrated in Figure~\ref{fig:motivation}, the framework first scores video segments using both mean query relevance and intra-segment relevance variance. It then applies a second sampling stage to refine the candidate segments most likely to contain critical evidence. The retained segments are subsequently organized into a temporal multimodal graph that encodes visual descriptions, transcripts, entities, relations, and event order. Finally, evidence is jointly retrieved from frame, text, and graph spaces, with temporal filtering applied when an explicit query time is available. In this way, VTG-RAG converts a long and redundant video stream into a compact, structured, and question-relevant evidence space for downstream reasoning.

We evaluate VTG-RAG on datasets including Video-MME and LongVideoBench, which cover multimodal video understanding and long-context video question answering, respectively~\cite{fu2024videomme,wu2024longvideobench}. Under matched visual-frame budgets, the framework produces accuracy improvements over the corresponding direct baselines across the evaluated vision--language backbones, including Qwen2.5-VL~\cite{bai2025qwen25vl} models ranging from 7B to 72B parameters. The efficiency measurements further show that the retrieved evidence requires fewer tokens than the corresponding direct-inference contexts. Component-wise ablations on Video-MME show that performance decreases when query-adaptive sampling, temporal graph retrieval, or two-pass reasoning is removed. Together, these results show that efficient sampling and evidence organization jointly improve long-video reasoning.

The main contributions of this work are as follows:
\begin{itemize}
  \item We present a selection-before-indexing pipeline in which
  query-adaptive video sampling determines the evidence used for
  multimodal graph construction.
  \item We connect the resulting graph with frame and text retrieval,
  followed by working-memory-guided evidence completion and
  answer generation.
  \item We report benchmark comparisons and grouped ablations that
  characterize the complete pipeline under the evaluated configurations,
  while distinguishing these observations from untested
  mechanism-level claims.
\end{itemize}

\section{Related Work}

\subsection{Video Retrieval-Augmented Generation}

Retrieval-augmented generation (RAG) is originally introduced for text-based, knowledge-intensive tasks. Its central idea is to explicitly retrieve external evidence to compensate for limitations in parametric memory and reduce hallucinations during generation~\cite{lewis2020rag}. With the development of multimodal large language models, this paradigm has been extended to video understanding. It addresses challenges in long-video settings, including excessively long contexts, dense visual tokens, and sparsely distributed critical evidence. MovieChat, MovieChat+, and LongVA improve long-video understanding through sparse memory modeling, question-aware compression, and long-context transfer, respectively~\cite{song2023moviechat,song2024moviechatplus,zhang2024longva}. VideoRAG and related studies further formulate long-video question answering explicitly as video retrieval-augmented generation. They support answer generation by constructing retrievable intermediate representations or directly retrieving relevant visual evidence from video corpora~\cite{jeong2025videorag,ren2025videorag}. However, most existing video RAG methods still use uniformly segmented clips or flat multimodal chunks as their basic indexing units. They neither screen critical temporal intervals before indexing nor explicitly model temporal structure. Consequently, these methods remain susceptible to noisy context in long-video question answering that requires sparse evidence localization, cross-segment aggregation, and temporally consistent reasoning.

\subsection{Graph-Based Visual Understanding and Temporal Reasoning}

Graph structures have been widely used in retrieval-augmented generation and visual understanding to represent complex relations and long-range dependencies that flat retrieval cannot capture. GraphRAG, LightRAG, and KG$^2$RAG organize dispersed evidence into structured retrieval spaces suitable for multi-hop reasoning. They use entity graphs, lightweight graph-enhanced indexing, and knowledge-graph-guided context expansion, respectively~\cite{edge2024graphrag,guo2024lightrag,zhu2025kg2rag}. In video understanding, EgoGraph further shows that temporal knowledge graphs can model entity interactions, event order, and state evolution across segments, thereby improving visual reasoning over long temporal spans~\cite{sun2026egograph}. These studies demonstrate that graph structures can strengthen relational representation and temporal modeling in complex visual reasoning. Existing approaches, however, generally assume that upstream video segments are already relatively clean. If a graph is constructed directly from many uniformly segmented and noisy video clips, its size can grow rapidly while its information density decreases, weakening the benefits of structured retrieval.

\subsection{Evidence Compression}

Efficiency improvements can target either model complexity or input redundancy.
Lu et al. investigate layer pruning and post-pruning fine-tuning to reduce
the computational overhead of large language models~\cite{lu2026reassessing}.
Unlike model compression, evidence compression reduces redundant input
while retaining task-relevant information for downstream reasoning.

For textual inputs, Gao et al. formulate prompt compression as
redundancy-aware pruning on a multiplex graph that captures local dependencies
and global semantic relations~\cite{gao2026mapping}.
Their method uses L\'evy walks to identify informative nodes, illustrating
a structure-aware approach to input compression.
For visual inputs, FOCUS formulates keyframe selection in long videos
as an upper-confidence-bound exploration problem, enabling training-free
frame selection under a constrained visual-token budget~\cite{zhu2025focus}.
FocusGraph further combines graph-structured scene representations with
hierarchical evidence selection.
It first identifies relevant clips and then refines keyframes within each clip
for embodied long-video question answering~\cite{zemskova2026focusgraph}.

These studies address complementary aspects of efficient inference and
evidence selection.
Our focus is on selecting sparse question-relevant video evidence and
organizing it for temporally consistent reasoning.
VTG-RAG therefore couples query-adaptive coarse-to-fine sampling with
temporal graph retrieval.
The retained evidence is organized by entities, relations, and temporal order
before multi-space retrieval and reasoning.

\section{Method}

\begin{figure}[t]
  \centering
  \includegraphics[width=\textwidth]{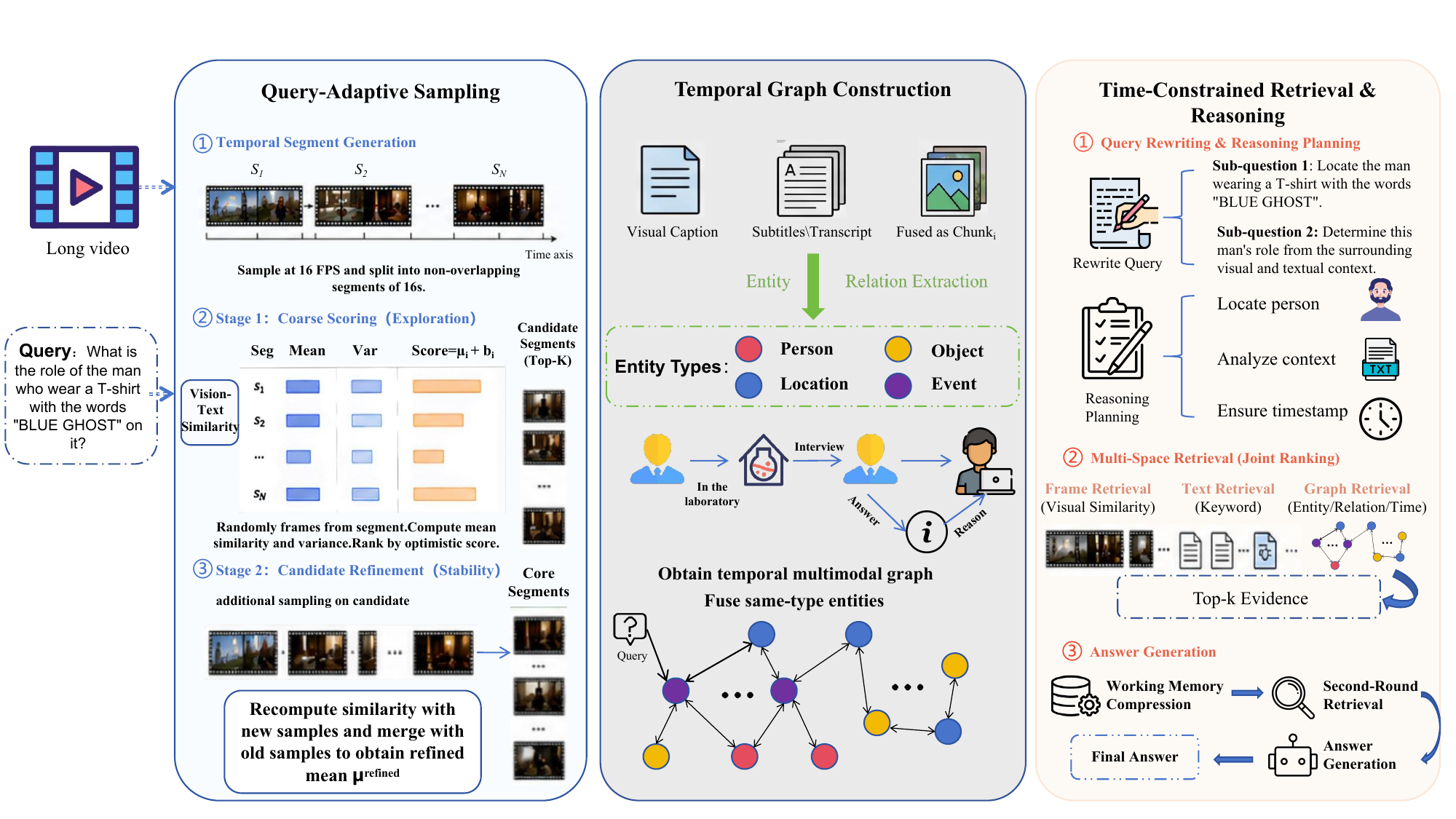}
  \caption{Overall architecture of VTG-RAG for long-video question answering. The framework comprises three modules. Query-Adaptive Coarse-to-Fine sampling combines variance-aware coarse scoring with candidate refinement to identify a compact set of core segments. Temporal graph construction integrates visual descriptions, transcripts, entities, relations, and temporal links into a multimodal graph. Time-constrained retrieval and reasoning combines query rewriting, reasoning planning, multi-space retrieval, and working memory to retrieve supporting evidence and generate the final answer}
  \label{fig:framework}
\end{figure}

Figure~\ref{fig:framework} presents the overall architecture of VTG-RAG. Given a long video $V$, a question $q$, and candidate answers $\mathcal{O}=\{o_1,\ldots,o_K\}$, where $K$ is the number of answer options, the objective is to generate a video-grounded answer $\hat{y}$. When a question time $t_q$ is available, retrieval is designed to exclude timestamped documents later than $t_q$, subject to the limitations described below. In Video-MME, time generally corresponds to the video playback timeline. For datasets that do not provide an explicit question time, such as LongVideoBench, the system retrieves evidence over the full video timeline. To this end, VTG-RAG follows three stages. First, query-adaptive sampling ranks temporal segments using their mean relevance and within-segment relevance variance, and then refines the top candidates into a compact set of core segments. Second, visual descriptions and transcripts from the retained segments are fused into multimodal chunks and organized as a temporal graph containing entities, relations, and event order. Finally, query rewriting and reasoning planning guide temporally constrained retrieval across frame, text, and graph spaces. The retrieved evidence is compressed into working memory and supplemented through a second retrieval pass before final answer generation. The central challenge is therefore to identify a compact yet sufficiently informative evidence set from a long, multimodal, and noisy video stream while preserving temporal consistency.

\subsection{Query-Adaptive Coarse-to-Fine Sampling}

Our sampler follows the two-stage optimistic selection principle of
FOCUS~\cite{zhu2025focus}. We use a variance-thresholded exploration
bonus to retain candidate segments, then resample these candidates
and rank them by their refined mean relevance. The bonus is a
heuristic adaptation of this selection principle, not a newly
derived statistical confidence bound. Its role here is to determine
which segments supply evidence for graph construction.

\textbf{Temporal Segment Generation.}
In the current implementation, the input video is partitioned into fixed-length, non-overlapping 16-second segments $\{s_1,\ldots,s_N\}$. From each segment $s_i$, the system randomly samples $k_c$ frames to form the sequence $F_i=\{f_{i,1},\ldots,f_{i,k_c}\}$. Each $F_i$ is the basic unit for subsequent scoring, representation, and graph construction.

\textbf{First-stage Coarse-grained Scoring.}
For each segment $s_i$, the system uses an image--text similarity model to score the relevance of the $k_c$ sampled frames in $F_i$ to question $q$. The current implementation uses a SigLIP-based local similarity scorer~\cite{zhai2023siglip}. Let the resulting similarity scores be $\{r_{i,1},\ldots,r_{i,k_c}\}$. The mean score of segment $i$ is defined as
\begin{equation}
\mu_i=\frac{1}{k_c}\sum_{j=1}^{k_c}r_{i,j},
\label{eq:mean}
\end{equation}
which measures the segment's overall match to the query. We further define the within-segment variance as
\begin{equation}
\sigma_i^2=\frac{1}{k_c}\sum_{j=1}^{k_c}(r_{i,j}-\mu_i)^2,
\label{eq:variance}
\end{equation}
which characterizes the variation in relevance within the segment. Unlike mean-only ranking, our approach also considers intra-segment relevance variance. A segment with moderate mean relevance may still contain a few highly relevant frames and therefore exhibit a large $\sigma_i^2$. Ranking only by $\mu_i$ may therefore discard segments containing decisive evidence at an early stage. We consequently define a variance-aware optimistic score as
\begin{equation}
\tilde{s}_i=\mu_i+b_i,
\label{eq:optimistic-score}
\end{equation}
where $b_i$ is a variance-dependent exploration bonus that compensates for the potential under-ranking of segments containing sparse but critical evidence. Intuitively, $b_i$ measures whether a segment merits retention for further examination. Even when its mean relevance $\mu_i$ is not among the highest, a segment may contain decisive evidence if a few sampled frames are highly relevant to the question. It should therefore receive an additional ranking bonus. Specifically, when the intra-segment variance $\sigma_i^2$ does not exceed the threshold $\tau$, we set
\begin{equation}
b_i=0;
\end{equation}
when $\sigma_i^2>\tau$, we define
\begin{equation}
b_i=\alpha\left(\sqrt{\frac{2\ln T\cdot\sigma_i^2}{n_i}}+\frac{3\ln T}{n_i}\right),
\end{equation}
Here, $n_i$ is the number of frames sampled from segment $i$, and $T=Nk_c$ is the nominal first-stage sampling budget for $N$ segments. The $\ln T$ factor adjusts the bonus according to the global sampling scale. The square-root term captures the principal contribution of relevance variance, whereas $\frac{3\ln T}{n_i}$ provides an additional conservative correction when the segment has been sparsely sampled. Thus, $b_i$ quantifies the extent to which a segment may be under-ranked by its mean score alone and allows the first stage to retain potentially critical segments for further examination. The coefficient $\alpha$ controls the strength of the bonus. For fixed $T$ and $n_i$, and conditional on $\sigma_i^2>\tau$, $b_i$ increases with $\sigma_i^2$. Low variance indicates relatively uniform relevance and warrants little or no bonus. High variance suggests that a few sampled frames may carry critical evidence and therefore yields a larger bonus. For fixed $T$ and $\sigma_i^2$, $b_i$ decreases as $n_i$ increases because denser sampling yields a more stable relevance estimate. In other words, $\sigma_i^2$ measures the within-segment heterogeneity associated with potentially critical evidence, and $b_i$ converts this heterogeneity into a ranking bonus. The system ultimately ranks all segments by $\tilde{s}_i$ and retains a candidate set subject to ratio and count constraints for subsequent second-stage refinement and retrieval-based reasoning.

\textbf{Second-stage Candidate Refinement.}

The second stage processes the candidate segments retained after ranking by the first-stage optimistic score. A second round of refinement sampling is performed only on these candidates. This design first retains potentially critical segments at relatively low cost and then concentrates additional computation on promising candidates, thereby controlling computational overhead.

The first-stage score is deliberately exploratory to avoid prematurely discarding potentially critical segments. However, the first stage estimates relevance from only a few sampled frames. It therefore favors recall and efficiency but may be sensitive to individual high-scoring frames. To obtain a more robust relevance estimate, VTG-RAG performs an additional round of sampling on the candidate segments retained after the first stage. Up to $m$ previously unsampled frames are drawn without replacement. Their relevance to the question is computed, and the new observations are combined with the first-stage samples to calculate a refined mean. Suppose that the first stage yields $k_c$ local similarity scores $\{r_{i,1},\ldots,r_{i,k_c}\}$ and that the second stage samples an additional $m$ frames with scores $\{r'_{i,1},\ldots,r'_{i,m}\}$. The refined mean is defined as
\begin{equation}
\hat{\mu}_i=\frac{1}{k_c+m}\left(\sum_{j=1}^{k_c}r_{i,j}+\sum_{t=1}^{m}r'_{i,t}\right).
\end{equation}
Unlike the first stage, the second stage no longer uses the optimistic score. Instead, it reranks the candidate segments according to the refined mean $\hat{\mu}_i$ and uses this ranking to determine the final set of core segments. The first stage preserves potentially relevant segments to reduce the risk of missing critical evidence. The second stage increases sampling density and reduces the influence of occasional high-scoring frames, producing the final core-segment ranking. Only the selected core segments proceed to visual-description generation, transcript alignment, temporal graph construction, and retrieval-based reasoning.

\subsection{Temporal Graph Construction}

Entity-centric temporal graphs are an established representation
for long-video understanding, as illustrated by
EgoGraph~\cite{sun2026egograph}. We do not claim Person, Location,
Object, and Event types or same-type entity fusion as independently
novel mechanisms. Here, these mechanisms organize the evidence
supplied by the preceding question-dependent segment selection stage.

\textbf{Multimodal Representation of Core Segments.}
For each retained segment, VTG-RAG extracts two types of textual evidence. First, it selects the $k_{\mathrm{cap}}$ sampled frames with the highest query-similarity scores and generates their visual descriptions using the configured local vision--language model. Second, it incorporates ASR-derived transcripts as speech evidence. For LongVideoBench, the textual evidence consists of these transcripts and generated visual descriptions; official subtitles are not used by VTG-RAG. The two sources are then fused into a segment-level chunk:
\begin{equation}
c_i=\operatorname{merge}(\operatorname{caption}_i,\operatorname{transcript}_i),
\label{eq:chunk}
\end{equation}
where each chunk stores the segment identifier, start and end times, and fused textual content. This representation preserves both what occurred and when it occurred and serves as the basic node unit for subsequent temporal graph construction.

\textbf{Chunk-level Temporal Backbone.}
After sorting the core chunks by start time, the system adds them as nodes in a directed multigraph. For each adjacent pair, it inserts a NEXT\_IN\_TIME edge from $c_i$ to $c_{i+1}$. This temporal backbone is essential for answering questions about event order, repeated mentions, and state transitions.

\textbf{Entity and Relation Extraction.}
For each chunk, the system extracts entities and relations from the fused descriptive text. In the current implementation, entity types are restricted to Person, Location, Object and Event. Each extracted entity may carry attribute information, and each relation is stored together with its supporting evidence and confidence score. When the remote extraction model is unavailable, the system falls back to a rule-based extractor to maintain pipeline robustness.

\textbf{Entity Fusion.}
The same person, location, object, or event may be mentioned repeatedly across different chunks. To reduce duplication and accumulate evidence over time, VTG-RAG fuses entities of the same type. The system first normalizes each entity to a canonical textual form and then computes embedding similarity. It merges a mention into an existing same-type node when their cosine similarity exceeds a threshold. The node then records the mention's timestamp. Otherwise, the system creates a new node. The resulting graph is converted into retrievable evidence comprising visual descriptions, subtitles or transcripts, descriptions of temporal order, entity nodes, edges, and relations.

\subsection{Time-constrained Retrieval and Reasoning}

\textbf{Query Rewriting and Reasoning Planning.}
Consider the original question: \textit{What is the role of the man who wears a T-shirt with the words ``BLUE GHOST'' on it?} VTG-RAG rewrites it for cross-modal retrieval as follows: \textit{Find the video segments showing the man wearing a T-shirt with the words ``BLUE GHOST''. Identify the surrounding scene, the man's actions and interactions, and any spoken or visual clues that indicate his role.} It then formulates the corresponding reasoning plan. First, the system retrieves segments showing the T-shirt bearing the words ``BLUE GHOST'' and locates the time range in which the man appears. Second, it uses the subtitles and visual descriptions to determine his actions, speech, and relationships with other people in the scene. Third, it retrieves entities, events, and adjacent segments associated with the man from the temporal graph to establish his identity or responsibility. Finally, it compares the evidence with the candidate answers and selects the role that is best supported. In the multiple-choice setting, the system also injects discriminative cues derived from the candidate answers to make subsequent evidence retrieval more sensitive to differences among the options. This process transforms the original question into a unified intermediate representation that supports both retrieval and reasoning, thereby providing a more explicit target for subsequent evidence aggregation.

\textbf{Multi-space Retrieval with Temporal Filtering.}
The retriever supports an optional task-supplied query-time cutoff
$t_q$. Candidate documents with known timestamps are filtered using
their end time, or their start time when no end time is available,
before ranking in each retrieval pass. Documents without timestamps
are not excluded by this rule.

When no cutoff is supplied, the video duration is used as the
temporal reference, leaving the full timeline eligible.
The temporal-proximity score then acts as a recency preference
rather than a causality constraint. LongVideoBench is evaluated
in this full-timeline setting.

Retrieval proceeds across frame descriptions, transcripts, and
graph-derived evidence, and the results are merged into a shared
ranking. No dedicated prefix-only evaluation is reported, so the
current results do not validate prevention of future-information
leakage. Filtering document timestamps alone cannot exclude
future-derived information already incorporated into fused
graph entities or summaries.

\textbf{Working-memory Compression and Second-pass Retrieval.}
Evidence obtained in the first retrieval pass is not sent directly to the answer-generation module. Instead, it is compressed into a compact working memory that retains the most relevant timeline summary, key facts, changes in entity states, and evidence for or against each candidate answer. Rather than answering in a single pass, VTG-RAG uses this working memory as an intermediate reasoning state to drive a second, complementary retrieval pass. This pass fills evidence gaps and suppresses distractors introduced during the initial retrieval. The answer model then generates the final prediction from the retrieved context, reasoning plan, and working memory, using timestamp-filtered retrieval when $t_q$ is available. This two-pass mechanism implements coarse recall followed by targeted evidence completion within the reasoning loop.

\section{Experiments}

\subsection{Datasets and Experimental Settings}

We evaluate VTG-RAG on two benchmarks: Video-MME~\cite{fu2024videomme} and LongVideoBench~\cite{wu2024longvideobench}. Video-MME results are reported separately with and without official subtitles. For LongVideoBench, we use the official labeled validation set containing 753 videos and 1,337 multiple-choice questions. Since no explicit query-time cutoff is provided, retrieval covers the full video timeline.

For Video-MME, the two subtitle conditions indicate whether
official subtitles are provided. ASR-derived transcripts remain
available to VTG-RAG in both conditions. Therefore, the setting
without official subtitles is not a visual-only evaluation.

For LongVideoBench, the direct baseline receives uniformly sampled
video frames and official subtitles, without ASR-derived transcripts
or generated visual descriptions. VTG-RAG uses ASR-derived
transcripts and generated visual descriptions for evidence
construction and retrieval. The paired configurations use the
same vision--language backbone and answer-input frame count,
but their textual inputs are not identical.

Videos are divided into non-overlapping 16-second segments.
A SigLIP-based image--text scorer supports query-adaptive
coarse-to-fine sampling. For LongVideoBench, each paired comparison uses the same
vision--language backbone and answer-input frame count.
The direct baseline receives uniformly sampled video frames
and official subtitles. VTG-RAG instead uses selected video
frames, ASR-derived transcripts, and generated visual descriptions,
without official subtitles. The two pipelines therefore use
different sources of textual evidence. Unless otherwise specified, the retrieved-text
budget is 2.0K tokens; configuration-specific budgets are
reported in Table~\ref{tab:longvideobench-results}.

\subsection{Baselines}

We compared VTG-RAG with LightRAG~\cite{guo2024lightrag}, VideoRAG~\cite{ren2025videorag}, KG$^2$RAG~\cite{zhu2025kg2rag}, and LongVA ~\cite{zhang2024longva}. 
\begin{itemize}[
  leftmargin=1.5em,
  topsep=3pt,
  itemsep=4pt,
  parsep=0pt,
  partopsep=0pt
]
\item 
KG$^2$RAG introduces knowledge graphs into retrieval-augmented generation. It extracts entities and relations from textual corpora, constructs a structured knowledge graph, and expands relevant evidence along the graph during retrieval. Although this organization benefits multi-hop question answering, the graph is designed primarily for static textual knowledge and does not explicitly model video-segment order or cross-segment event evolution.
\item 
LongVA transfers the long-context capability of language models to visual understanding. It processes large numbers of video frames or visual tokens to model long videos end to end, providing direct access to information distributed over long temporal spans. However, it relies primarily on a large visual context for global reasoning and does not actively select question-relevant segments or construct structured evidence before inference.
\item 
VideoRAG combines static knowledge graphs with video content. It is developed for lecture videos, which generally exhibit less visual variation than egocentric videos, where head motion can produce substantial viewpoint changes. Its applicability to extremely long egocentric videos may therefore be limited.
\item 
LightRAG is a text-based retrieval-augmented generation method. It divides a long context into chunks and extracts nodes and edges from each chunk. This graph-based retrieval approach is particularly suitable for query-focused summarization. Our setting follows a related idea by tracking recurring entities across temporally separated segments in extremely long videos.
\end{itemize}

Under the configurations considered here, these methods do not provide the explicit temporal modeling targeted by VTG-RAG. In contrast, VTG-RAG constructs a temporally ordered multimodal graph for explicit temporal reasoning. Unless otherwise specified, SigLIP2 is used as the embedding model for all retrieval-based comparison methods to ensure a fair comparison. We also use the same question-answering model, Qwen2.5-VL-7B, for all retrieval-based methods.

\begin{table}[t]
\centering
\caption{Comparison of the main results on Video-MME. $\Delta_{\mathrm{Ours}}$ denotes the absolute improvement of VTG-RAG in Avg over the corresponding method. The subtitle conditions refer to the availability of official
subtitles. VTG-RAG uses ASR-derived transcripts in both conditions; the condition without official subtitles is therefore not visual-only.}
\label{tab:main-results}

\begin{tabular*}{\textwidth}{@{\extracolsep{\fill}}lccccc@{}}
\toprule
\multirow{2}{*}{Model} & \multirow{2}{*}{Size} & \multicolumn{2}{c}{Video-MME} & \multirow{2}{*}{Avg} & \multirow{2}{*}{$\Delta_{\mathrm{Ours}}$} \\
\cmidrule(lr){3-4}
& & \shortstack{w/o official\\subtitles} & \shortstack{w/ official\\subtitles} & & \\
\midrule
\multicolumn{6}{c}{\textit{Proprietary LVLMs}} \\
\midrule
Gemini 1.5 Pro & -- & 42.69 & 48.46 & 45.58 & -- \\
GPT-4o & -- & 53.62 & 59.17 & 56.40 & -- \\
\midrule
\multicolumn{6}{c}{\textit{Open-source LVLMs}} \\
\midrule
LLaVA-OV & 7B & 35.08 & 35.52 & 35.30 & -- \\
LongVA & 7B & 30.16 & 33.29 & 31.72 & -- \\
InternVL3 & 8B & 28.94 & 29.82 & 29.38 & -- \\
\midrule
 \multicolumn{6}{c}{\textit{Graph-based methods}} \\
\midrule
LightRAG & -- & 37.80 & 40.78 & 39.29 & 3.05 \\
KG$^2$RAG & -- & 32.45 & 32.16 & 32.31 & 10.03 \\
VideoRAG & -- & 33.77 & 33.77 & 33.77 & 8.57 \\
\textbf{Ours (Qwen2.5-VL)} & 7B & \textbf{40.78} & \textbf{43.89} & \textbf{42.34} & / \\
\bottomrule
\end{tabular*}
\end{table}

\textbf{Analysis of The Main Results.}
Table~\ref{tab:main-results} shows that VTG-RAG obtained an Avg score of 42.34, higher than the three graph-based baselines shown. Because the systems differ in multiple components, this comparison alone does not isolate the effects of query-adaptive coarse-to-fine sampling or temporal graph retrieval. Table~\ref{tab:longvideobench-results} presents the performance and efficiency of VTG-RAG with different vision--language backbones on LongVideoBench. For each backbone, we match the model parameter count and
answer-input frame count. The direct baseline uses uniformly sampled frames and official subtitles, whereas VTG-RAG uses selected frames, ASR-derived transcripts, and generated visual descriptions without official subtitles. This comparison therefore evaluates complete pipelines with different textual evidence sources. With LongVA, LLaVA, InternVL3, and Qwen2.5-VL, VTG-RAG yields absolute accuracy gains of 15.9, 16.4, 13.6, and 11.1 percentage points, respectively. These gains are observed across the tested architectures, parameter counts, and visual-frame budgets, although the available results do not establish generalization beyond these settings.

VTG-RAG uses only 1.8K--2.1K tokens of retrieved textual evidence and achieved higher accuracy under a fixed visual-frame budget. Under the evaluated input configurations, VTG-RAG uses fewer reported tokens than the corresponding direct baselines. Across the four completed settings, reported token use decreases from 11.3M--21.6M to 4.8M--10.4M, an average reduction of approximately 50\%. Correspondingly, accuracy per 1M tokens increases from 1.45--1.83 to 3.86--7.33 percentage points per million tokens. Thus, the reported settings show concurrent improvements in answer accuracy and token-use efficiency.

The number of frames affects baseline performance, but more frames do not necessarily yield higher accuracy. For example, the Qwen2.5-VL direct baseline achieves the highest accuracy with 24 frames. LongVA, LLaVA, and InternVL3 still differ substantially when using 8 or 16 frames. These results indicate that frame count alone does not determine performance within the evaluated settings. The video-understanding capability of the backbone and its use of evidence may also have contributed. The paired settings use the same number of answer-input frames, so the accuracy gains do not arise from increasing this frame count. However, the textual evidence sources differ between the pipelines. These results therefore support the evaluated system configurations, but do not isolate the contributions of query-adaptive sampling or temporal graph retrieval.

Table~\ref{tab:longvideobench-results} reports the comparisons using 7B and 8B backbones. Results for larger Qwen2.5-VL models are reported in Section~\ref{sec:scaling}. Within these settings, VTG-RAG improves Qwen2.5-VL from 34.9\% to 46.0\% and InternVL3-8B from 23.2\% to 36.8\%. These observations suggest that Query-Adaptive Coarse-to-Fine sampling and temporal graph retrieval may complement the visual and language modeling capacity of the tested backbones.

\begin{table}[t]
\centering
\caption{Performance and token efficiency of different vision--language models on LongVideoBench. Text denotes the retrieved-text budget of VTG-RAG, and Gain denotes the absolute accuracy improvement over the direct baseline using the same model. Direct baselines use official subtitles, whereas VTG-RAG uses ASR-derived transcripts and generated visual descriptions without official subtitles. The Text column reports the retrieved-text budget of VTG-RAG; ``--'' does not indicate the absence of subtitle input for the direct baselines}
\label{tab:longvideobench-results}
\small
\setlength{\tabcolsep}{3pt}
\begin{tabular*}{\textwidth}{@{\extracolsep{\fill}}lccccccc@{}}
\toprule
Model & Text & Size & Frames & \shortstack{Acc.\\(\%)} & \shortstack{Gain\\(pp)} & Tokens &
\shortstack{Acc./1M\\tokens} \\
\midrule
\multicolumn{8}{c}{\textit{LVLMs}} \\
\midrule
LongVA & -- & 7B & 8 & 24.2 & -- & 16.7M & 1.45 \\
LongVA + VTG-RAG & 1.8K & 7B & 8 & 40.1 & +15.9 & 10.4M & 3.86 \\
\midrule
LLaVA & -- & 7B & 16 & 18.8 & -- & 11.3M & 1.65 \\
LLaVA + VTG-RAG & 2.0K & 7B & 16 & 35.2 & +16.4 & 4.8M & 7.33 \\
\midrule
InternVL3 & -- & 8B & 16 & 23.2 & -- & 12.7M & 1.83 \\
InternVL3-8B + VTG-RAG & 2.0K & 8B & 16 & 36.8 & +13.6 & 6.8M & 5.41 \\
\midrule
Qwen2.5-VL & -- & 7B & 24 & 34.9 & -- & 21.6M & 1.62 \\
Qwen2.5-VL + VTG-RAG & 2.1K & 7B & 24 & 46.0 & +11.1 & 8.9M & 5.17 \\
\bottomrule
\end{tabular*}
\end{table}

\subsection{Scaling to Larger Models}\label{sec:scaling}

On the reported LongVideoBench configurations with 24 answer-input frames, Qwen2.5-VL-32B improves from 48.16\% to 50.56\%, a gain of 2.40 percentage points. Qwen2.5-VL-72B improves from 52.13\% to 53.32\%, a gain of 1.19 percentage points.

The larger model attains higher absolute accuracy, but its gain from VTG-RAG is smaller. These two configurations do not establish a monotonic relationship between model size and pipeline benefit. Statistical significance and robustness to sampling randomness remain to be established.

\begin{table}[t]
\centering
\caption{Scaling performance of VTG-RAG with larger Qwen2.5-VL models on the official labeled validation set of LongVideoBench. All configurations used 24 visual frames and the same evaluation protocol. Retrieved text denotes the textual-evidence budget. Gain is the absolute accuracy improvement, in percentage points, over the direct baseline with the same model size and visual-frame budget}
\label{tab:large-model-scaling}
\small
\setlength{\tabcolsep}{3pt}
\begin{tabular*}{\textwidth}{@{\extracolsep{\fill}}lccccc@{}}
\toprule
Model & \shortstack{Retrieved\\text} & Size & Frames & \shortstack{Accuracy\\(\%)} & \shortstack{Gain\\(pp)} \\
\midrule
Qwen2.5-VL
    & --   & 32B & 24 & 48.16 & -- \\
Qwen2.5-VL + VTG-RAG
    & 2.1K & 32B & 24 & \textbf{50.56} & \textbf{+2.40} \\
\midrule
Qwen2.5-VL
    & --   & 72B & 24 & 52.13 & -- \\
Qwen2.5-VL + VTG-RAG
    & 2.1K & 72B & 24 & \textbf{53.32} & \textbf{+1.19} \\
\bottomrule
\end{tabular*}
\end{table}

\subsection{Ablation Experiments}

We conduct three grouped ablations to examine the contributions of query-adaptive sampling, graph retrieval, and the two-pass reasoning pipeline. Table~\ref{tab:ablation} reports the results on Video-MME Long with subtitles. These variants evaluate module-level changes rather than isolate every operation within each module.

\begin{table}[t]
\caption{Core ablation experiments for VTG-RAG. All results are reported under the Video-MME Long setting with subtitles}
\label{tab:ablation}
\centering
\begin{tabular*}{\columnwidth}{@{\extracolsep{\fill}}lc@{}}
\toprule
Configuration & Long w/ subtitles (\%) \\
\midrule
Full model & \textbf{43.8} \\
\quad -- w/o query-adaptive sampling & 40.7 \\
\quad -- w/o graph retrieval & 42.1 \\
\quad -- w/o two-pass reasoning & 42.9 \\
\bottomrule
\end{tabular*}
\end{table}

\textbf{Role of Query-Adaptive Sampling.}
 To evaluate the contribution of Query-Adaptive sampling, we construct a uniformly sampled variant. This variant replaces Query-Adaptive sampling in the first stage with uniform segment selection while retaining all other modules. Compared with the full model, its accuracy on Video-MME decreases from 43.8 to 40.7. This decrease is consistent with question-guided segment retention helping allocate computation to relevant context and retain key evidence.

\textbf{Role of Temporal Graph Retrieval.}
To examine temporal graph retrieval, we construct a variant without graph retrieval. This variant retains frame-level visual evidence and textual transcripts but omits graph retrieval based on entity relations and temporal dependencies. Removing the temporal graph reduces performance on Video-MME to 42.1. This degradation is consistent with structured temporal information contributing to the aggregate performance of the full model.

\textbf{Role of Two-pass Reasoning.}
To assess the contribution of two-pass reasoning, we further remove working-memory compression and second-pass retrieval. The model instead generates an answer directly from the first-pass retrieval results. This variant achieves 42.9 on Video-MME, compared with 43.8 for the full model. The result is consistent with working-memory compression and second-pass retrieval helping organize initial evidence and suppress redundant or distracting information.

\begin{figure}[tbp]
\centering
\includegraphics[width=0.78\textwidth]{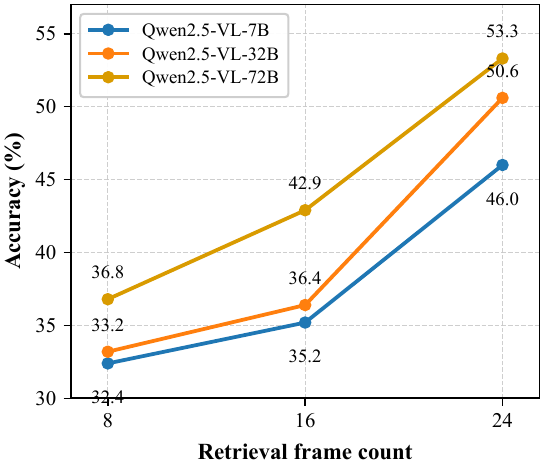}
\caption{Effect of the retrieval frame count on Qwen2.5-VL models at different scales}
\label{fig:rateline}
\end{figure}

\subsection{Effect of Retrieval Frame Count Across Model Scales}
To investigate how the retrieval frame count affects models of different scales, we evaluate Qwen2.5-VL-7B, Qwen2.5-VL-32B, and Qwen2.5-VL-72B using frame budgets of $F \in \{8,16,24\}$, while keeping all other settings fixed. As shown in Figure~\ref{fig:rateline}, increasing the number of retrieved frames consistently improves accuracy across all three model scales. Specifically, the accuracy of Qwen2.5-VL-7B increases from 32.4\% at $F=8$ to 35.2\% at $F=16$ and 46.0\% at $F=24$. Similar trends are observed for Qwen2.5-VL-32B, whose accuracy rises from 33.2\% to 36.4\% and 50.6\%, and for Qwen2.5-VL-72B, which improves from 36.8\% to 42.9\% and 53.3\%. Overall, increasing the frame budget from 8 to 24 yields absolute gains of 13.6, 17.4, and 16.5 percentage points for the 7B, 32B, and 72B models, respectively. Larger models also consistently outperform smaller models under the same frame budget, indicating that increased model capacity facilitates more effective use of the retrieved visual evidence. The largest performance improvements generally occur when the frame count increases from 16 to 24, suggesting that smaller frame budgets may omit evidence required for long-video reasoning. Therefore, we use 24 retrieved frames in the main experiments, as this setting achieves the best performance among the evaluated frame budgets.

\section{Discussion}

The central contribution of VTG-RAG is the coupling of query-adaptive evidence compression with structured temporal retrieval. The reported gains suggest that long-video question answering may depend on more than the strength of the retriever or answer model. Performance may also depend on actively controlling which video regions enter the reasoning pipeline.

The framework also embodies a design principle that may be useful for long-video RAG: temporal structure should be represented in the evidence structure and enforced during retrieval. VTG-RAG implements this principle at two levels. First, temporal information is encoded explicitly in the graph through chunk timestamps and NEXT\_IN\_TIME edges. Second, timestamped documents are filtered during retrieval using the timestamp $\leq t_q$ criterion. As noted above, this document-level filter does not guarantee that fused graph evidence is free from future-derived information. This design could be useful for question-answering settings with causal boundaries, including EgoLifeQA and other settings in which an explicit question time is available~\cite{yang2025egolife}.

The LongVideoBench comparison does not isolate the contribution of additional ASR and caption-derived evidence from that of evidence selection and organization.

The current system nevertheless has several limitations. First, graph quality depends directly on the quality of visual captions and entity--relation extraction. Weak captions or noisy extraction can make the graph sparse or even mislead retrieval. Second, the current graph schema is deliberately lightweight, which facilitates reproduction but provides limited support for abstract representations of complex events. Third, front-end sampling relies primarily on image--text similarity. It may therefore miss a critical segment when the decisive evidence lies mainly in linguistic detail rather than visual change. Future work could investigate stronger event-centric graph schemas, uncertainty-aware entity fusion, finer-grained subtitle--vision alignment, and adaptive segment lengths.

Another direction is to reduce the cost of annotating and validating
textual evidence used in graph construction.
AutoAnnotator coordinates large and small language models for efficient
text annotation~\cite{lu2025autoannotator}.
Adapting this collaborative strategy to video-derived descriptions and
entity--relation annotations remains a direction for future investigation.

\section{Conclusion}

We introduce VTG-RAG, a framework that combines query-adaptive coarse-to-fine sampling and temporal graph retrieval for long-video question answering. The central idea is to compress a long video into a small set of question-relevant core segments before graph construction. The system then jointly retrieves frame, text, and graph evidence, applying an explicit temporal filter when a question time is available. This design retains sparse but important evidence while reducing irrelevant context during downstream reasoning. First, we develop a query-adaptive coarse-to-fine sampling strategy. It combines segment--question relevance with within-segment variance in relevance scores to locate informative segments along a long timeline and compress redundant context. Second, we construct a temporally ordered multimodal graph with entity relations. The graph unifies visual descriptions, subtitles or transcripts, and entity relations within a structured evidence space to support cross-segment evidence integration. Third, we design time-constrained multi-space retrieval and two-pass reasoning. Under the stated conditions, this process jointly retrieves frame, text, and graph evidence and uses working memory for evidence completion and answer inference. Across the evaluated Video-MME and LongVideoBench settings, VTG-RAG improves accuracy relative to the selected graph-based or paired direct baselines. The completed LongVideoBench settings also show higher token efficiency. These findings are limited to the reported datasets, backbones, and frame budgets; statistical significance remains to be established.

\backmatter
\section*{Statements and Declarations}
\subsection*{Funding}
This work was supported by the National Key R\&D Program of China (2025YFA1510900), the Baima Lake Laboratory Joint Fund of the Zhejiang Provincial Natural Science Foundation of China (LBMHZ25F020002), the Yangtze River Delta Science and Technology Innovation Community Joint Research Project (2025CSJGG01000 and 2026ZY03003), the Applied Basic Research Program of Zhejiang Province (2026C02A1264), and the National Natural Science Foundation of China (62673439).

\subsection*{Competing interests}
The authors declare that they have no known competing financial interests or personal relationships that could have appeared to influence the work reported in this paper.

\subsection*{Author contributions}
\textbf{Yao Lu}: Conceptualization, Methodology, Software, Writing - original draft, Visualization. \textbf{Yunhao Wang}: Conceptualization, Methodology, Writing - review \& editing. \textbf{Yaxin Gao}: Validation, Formal analysis, Investigation. \textbf{Feifei Teng}: Conceptualization, Writing - original draft, Visualization. \textbf{Chenxiao Liu}: Conceptualization, Visualization. \textbf{Zhe Tang}: Investigation, Data curation. \textbf{Jiaheng Wei}: Investigation, Visualization. \textbf{Zhaowei Zhu}: Visualization, Writing - review \& editing, Supervision. \textbf{Shanqing Yu}: Resources, Writing - review \& editing, Supervision. \textbf{Qi Xuan}: Resources, Funding acquisition.

\subsection*{Data availability}
The study uses the Video-MME and LongVideoBench benchmarks described in Refs.~\cite{fu2024videomme,wu2024longvideobench}. The benchmark data and access instructions are available through the official project repositories at \url{https://github.com/MME-Benchmarks/Video-MME} and \url{https://github.com/longvideobench/LongVideoBench}, respectively. Access and reuse are subject to the conditions and licences specified by the original providers.

\subsection*{Code availability}
The implementation is not currently deposited in a public repository. The code required to reproduce the reported experiments is available from the corresponding author upon reasonable request to support verification and reproducibility.

\bibliography{references}
\end{document}